\documentclass[a4paper]{spie}  %>>> use for US letter paper
\usepackage{graphicx}
\usepackage{amssymb, amsmath, subfigure}
\providecommand{\norm}[1]{\lvert#1\rvert}
\title{Optimizing genetic algorithm strategies for evolving networks} 
\author{Matthew J. Berryman\supit{a}, Andrew Allison\supit{a}, and Derek Abbott\supit{a}
\skiplinehalf
\supit{a}Centre for Biomedical Engineering and\\
School of Electrical and Electronic Engineering,\\
The University of Adelaide, SA  5005, Australia.}
\authorinfo{Send correspondence to Derek Abbott\\
E-mail: dabbott@eleceng.adelaide.edu.au, Telephone: +61 8 8303 5748}
\begin{document} 
\maketitle
\begin{abstract}
This paper explores the use of genetic algorithms for the design of networks, where the demands on the network fluctuate in time. For varying network constraints, we find the best network using the standard genetic algorithm operators such as inversion, mutation and crossover. We also examine how the choice of genetic algorithm operators affects the quality of the best network found. Such networks typically contain redundancy in servers, where several servers perform the same task and pleiotropy, where servers perform multiple tasks. We explore this trade-off between pleiotropy versus redundancy on the cost versus reliability as a measure of the quality of the network.
\end{abstract}
\section{INTRODUCTION}
Evolutionary computation uses solution space search procedures inspired by biological evolution~\cite{WinstonAI}. These search procedures use ideas from biological evolution such as mating, fitness, and natural selection. Individuals undergo natural selection, whereby organisms with the most favorable traits are more successful in having offspring. Genetic algorithms (GAs) rely on describing systems in terms of their traits (or phenotype) and then a fitness function (or how well they reproduce). Then we can evolve better solutions, with a higher fitness function, by allowing transfer of hereditary characteristics (genes) to the next generation for fit functions. The idea of applying such biological concepts to evolutionary computing originated with John Holland in his seminal paper on the topic of adaptive systems~\cite{JohnHolland}.

Evolutionary computational techniques such as genetic algorithms have many advantages over traditional optimization algorithms. Current optimization algorithms require many assumptions to be made about the problem, for example with gradient-based searches, the requirement is that the function be smooth and differentiable. Evolutionary algorithms require no such assumptions, only requiring a way of measuring the ``fitness'' of a solution~\cite{IntroGA}. With each succeeding generation, the algorithm tries to better fulfill the specifications described by the fitness function. The other advantage is adaptability to a changing problem. For example with traditional optimization procedures, any change in the specification or problem constraints requires solving the problem from the start. This is not necessary with evolutionary algorithms where one can continue the algorithm with a different set of constraints or solution using the current ``population'' or set of solutions~\cite{Fogel}. GAs can offer advantages over related techniques such as hill climbing~\cite{mitchell94when,mitchell97}. Of great importance is the requirement of the genetic algorithm to support mating, or crossover between two graphs~\cite{mitchell94when}. In our previous work~\cite{Berryman3}, we only considered mutation operators. Thus our solution was never able to truly explore a wide region of the search space, and apart from initial hill climbing and other wide random jumps was quite slow at improving the network design. This current paper details the existing background to our work and the ongoing work of implementing crossover operators and different fitness functions.

Although there are a large number of applications of genetic algorithms to designing neural networks~\cite{neuralnet1,neuralnet2,neuralnet3}, there are very few devoted to designing computer or telecommunication networks~\cite{netga1,netga2}, and none of these explicitly capture the issue of pleiotropy. An alternative evolutionary approach using cellular automata has been applied to pleiotropy versus redundancy tradeoffs in an organizational system~\cite{Teck}. Pleiotropy is a term used to describe components that perform multiple tasks~\cite{pleiotropy1,pleiotropy2}, while redundancy refers to multiple components performing one same task. Such pleiotropy and redundancy of components can be clearly seen in a client-server based network comprising of server nodes and client nodes. The conventional setup of such networks can have servers serving multiple clients, which is an example of pleiotropy, while clients can be connected to many servers, which is an example of redundancy. 

A typical engineering problem is to determine the optimal design solution or set of solutions, while maximizing efficiency. The main aim of the project is to use evolutionary computation algorithms to search for an optimal client-server network, which minimizes cost and maximizes reliability and flexibility by exploring the pleiotropy-redundancy search space. 

\begin{figure}[hptb]
  \centering{\resizebox{6cm}{!}{\includegraphics{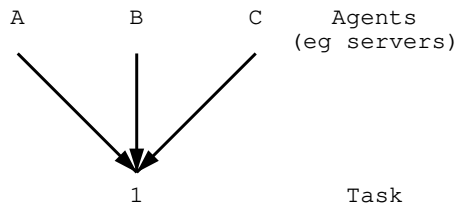}}}
  \caption{This figure shows the task, labeled 1, being performed by agents A, B, and C; thus two of these agents are redundant. An example in a network situation would be load-balancing a web server, where any one of three servers can serve a particular site to a client. Note that there is an extra cost associated with this redundancy, so in this example we are paying for three servers instead of one. However the system is robust, so if one of the servers is busy or breaks down, then the task (such as serving a web site) can still be performed.}
  \label{redundancy}
\end{figure}

Redundancy is where one task or outcome is carried out by more than one agent -- assuming the independence of agents, either acting independently or together.  It has the advantage of conferring robustness or integrity upon the system, because if one agent were to fail, others are able to perform the task.  However, redundancy may be costly, as the overlapping of agents may be inefficient or wasteful.  Despite this, in some systems the wastage may be justified if the task or outcome is so important that the system will fail in its absence, and therefore be selectively disadvantaged. Figure~\ref{redundancy} shows an example of a redundant system.

The opposite of redundancy is pleiotropy, where one agent may perform many tasks.  This has a number of distinct advantages.  It is efficient, and allows for spatial and temporal flexibility.  Its major cost is that it is dependent upon the history of the system, that is, any given agent may only be working under certain constraints imposed upon it by the peculiar evolutionary history of that system.  Despite this, both temporal and spatial pleiotropy may exist, where a given agent can perform qualitatively different tasks, as well as perform the same task (or different tasks) at different times.  As pleiotropy enables efficiency, it therefore confers selective advantages.  How pleoitropic an agent is will depend upon the context in which it is working.  
\begin{figure}[hptb]
  \centering{\resizebox{6cm}{!}{\includegraphics{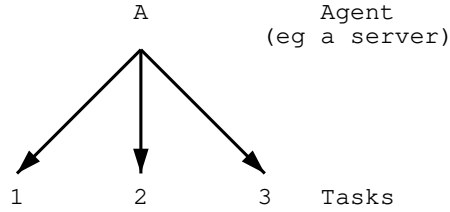}}}
  \caption{This figure shows a single agent, A, performing multiple tasks, labeled 1, 2 and 3. An example in a network situation could be a server handling multiple client requests, such as sending email to client 1 while sending a web page to client 2. While this is cost effective, it lacks the robustness to failure of a redundant system as shown in Figure~\ref{redundancy}.}
  \label{pleiotropy}
\end{figure}

\begin{figure}[hptb]
  \centering{\resizebox{8cm}{!}{\includegraphics{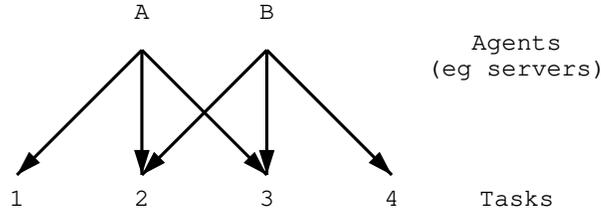}}}
  \caption{This figure shows two agents, both of which perform multiple tasks. An example in a network situation would be two servers, both providing the same email and web services to a number of clients. This system has robustness as if one server fails, both email and web services can still be provided. It also minimizes cost, as it would cost the same as two servers with one providing just email services and the other hosting a web site.}
  \label{both}
\end{figure}

What happens when you combine the two? When both pleiotropy and redundancy are combined, the system possesses properties that it otherwise lacks when pleiotropy or redundancy exist on their own.  The advantages include an increase in the robustness of the system due to the redundancy build into it.  It is more efficient due to the pleiotropy.  The system becomes inherently more flexible and the costs of redundancy are offset by the increase in efficiency due to the presence of pleiotropy. An example of a system with a combination of pleiotropy and redundancy is shown in Figure~\ref{both}.
\section{METHODS}
In this section we describe the structure and representation of the network, the details of the genetic algorithm used, the initialization and parameters used for designing the network using the genetic algorithm, and the fitness function. We developed a graphical user interface (GUI) for running the genetic algorithm, to allow for easy user modification of the network parameters, this also allows one to watch the evolution of the network.
\subsection{Network structure}
The network consists of a set of servers, a set of clients (which can also function as routers), and a set of links between those various nodes. A graph data structure is used to represent the network, with each node (client or server) in the graph having the following properties:
\begin{itemize}
\item node label, ``C'' for a client (including routers) or ``S'' for a server
\item node ID, which also serves as a grid reference of the node for display in a GUI
\item node failure rate, a value between zero and one giving the probability of failure per time step
\item current state, working or non-working
\item number of time steps since failure, zero if working
\item details of the inbound and outbound network connections.
\end{itemize}
The edges, the links in the network, have the following properties:
\begin{itemize}
\item link label indicating whether the link is a link between clients (including routers) or between a client or router to a server
\item edge ID, which also serves as a pair of grid references for display in a GUI
\item edge failure rate, a value between zero and one giving the probability of failure
\item current state, working or non-working
\item number of time steps since failure, zero if working.
\end{itemize}
We note here that for the set of nodes in a network, $\mathcal{N}$, the set of edges $\mathcal{E}$ is 
$\mathcal{E} \subseteq \mathcal{N} \times \mathcal{N}$. This becomes important where we later consider the crossover operation.
\subsection{Network construction}
We initially start with a set of clients ($\mathcal{C}$) and servers ($\mathcal{S}$), with no links. The positions of the clients and servers are set at random, with a minimum spacing between them.
Each client $i \in \mathcal{C}$ is assigned a traffic value, $T_{i}$, at random ($0<T_{i}<T_{\mathrm{max}}$), which indicates the amount of traffic requested by the client that is to be transmitted across the network. Each server $j \in \mathcal{S}$ has a fixed amount $T_{s}$ of traffic it can serve. 
\subsection{Mutation operator}
We define a utilization parameter,
\begin{equation}
U=\frac{\displaystyle \sum_{i \in \mathcal{C}} T_{i}}{\norm{\mathcal{S}}T_{s}},
\label{utilization}
\end{equation}
where $\norm{\mathcal{S}}$ is the size of the set of servers, describing how well the servers are able to deliver their available load to the clients.
If the utilization is less than $0.75$, then more links are added at random to carry the extra server capacity to clients. If, on the other hand, the utilization is greater than $0.85$ then either links are removed (reducing the amount of traffic that is able to be requested from servers) or more servers are added. The network thus evolves by starting without any connections, and through mutations including:
\begin{itemize}
\item adding links to increase $U$
\item removing links to decrease $U$
\item adding servers to decrease $U$
\item links failing
\item links being repaired.
\end{itemize}
\subsection{Crossover operator}
Consider two graphs $a$ and $b$, with nodes $\mathcal{N}_{a}$ and $\mathcal{N}_{b}$, and edges $\mathcal{E}_{a} \subseteq \mathcal{N}_{a}^{2}$ and $\mathcal{E}_{b} \subseteq \mathcal{N}_{b}^{2}$ respectively. We then wish to consider forming a new graph $c$, with a set of edges $\mathcal{E}_{c}\subseteq \mathcal{E}_{a}\cup \mathcal{E}_{b}$. 
To do this, we need to make sure that each node used by $\mathcal{E}_{c}$ appears in $\mathcal{N}_{c}$, since we require $\mathcal{E}_{c}\subseteq \mathcal{N}_{c}^{2}$. We could add each node associated with $\mathcal{E}_{c}$ to $\mathcal{N}_{c}$, guaranteeing that $\mathcal{E}_{c} = \mathcal{N}_{c}^{2}$, in $O(MN)$ time (for $M$ edges and $N$ nodes). Instead, we simply create the set of nodes $\mathcal{N}_{c}=\mathcal{N}_{a}\cup\mathcal{N}_{b}$ in $O(N)$ time. This (in the average case) overestimates the number of nodes required, somewhat breaking the role of the repair rate of the mutation operator. However, considering this is not clearly defined for crossover anyway, we choose the faster solution. For a selection of the top $q$ networks for mating, we have a total population size of $\left(q^{2}-q\right)/2+q$. The default value of $q$ we use is $q=5$, giving a total population size of $15$, however we vary this to ascertain the effect on convergence on the optimal solution.

Having established a network and a set of genetic algorithm operators, we then need a measure of fitness, in order to define that we first introduce Dijkstra's algorithm.
\subsection{Dijkstra's algorithm}
Dijkstra's algorithm is an efficient algorithm for finding the shortest path between two nodes (or vertices) in a graph~\cite{Dijkstra,shortPath}. Define $p_{i,j}$ to be the shortest path from node $i$ to node $j$. If there is an edge $e=(k,j)$ on $p_{i,j}$ then we can break down the path into $p_{i,k}$ and $(k,j)$. The distance, $d_{i,j}^{m}$ of the shortest path from $i$ to $j$, including at most $m$ edges, can be written as:
\begin{equation}
d_{i,j}^{(k)}=\min \left(d_{i,j}^{k-1},\underset{1\leq m \leq n}{\min}\left(d_{i,k}^{(m-1)}+w\left(e\right)\right)\right),
\label{d1}
\end{equation}
for $w\left(e\right)$ the edge cost of $e=\left(k,j\right)$ and $\left\{1,\ldots,n\right\}$ the set of nodes. What we wish to calculate is the cost $d_{i,j}^{(n)}$ for all pairs of nodes $i$ and $j$. By starting with the adjacency matrix $\pmb{A}=\left[w\left(i,j\right)\right]$ where $w\left(i,j\right)=\infty$ for $\left(i,j\right)$ not an edge, we compute the shortest distance matrix $\pmb{D}^{(n-1)}$ by calculating $\pmb{A}^{2},\pmb{A}^{4},\ldots,\pmb{A}^{n-1}$ in $\lceil \log\left(n-1\right) \rceil$ modified matrix multiplications where $\times$ is replaced by $+$ and $+$ is replaced by $\min$ in the computations of each element. Note that $d_{i,j}^{(n)}=\infty$ where no path exists between $i$ and $j$, so it provides information on connectivity in addition to costs. 
\subsection{Fitness and cost functions}
The aim is to find an optimal network, which minimizes cost ($P$) and maximizes reliability ($R$). With this in mind, we define our fitness function, $F$, to be:
\begin{equation}
F=\frac{R}{P},
\label{fitness}
\end{equation}
for a cost function $P$ and a reliability function $R$. Our cost function is given by
\begin{equation}
P=
\frac{\displaystyle\sum_{\left\{\left(i,j\right)\vert w\left(i,j\right)<\infty\right\}}w\left(i,j\right)}{\displaystyle\sum_{\left\{\left(i,j\right)\vert d_{i,j}^{(n)}<\infty\right\}}d_{i,j}^{(n)}},
\label{P}
\end{equation}
for $w\left( \cdot \right)$ and $d_{i,j}^{(n)}$ as defined above. Minimizing this cost function decreases the total cost of the links towards the minimum cost graph for that connectivity. Using just the cost alone results in a system that tends towards no links over many generations. Our reliability function is defined by 
\begin{equation}
R=\displaystyle\sum_{\left\{\left(i,j\right)\vert d_{i,j}^{(n)}<\infty\right\}}1,
\label{R}
\end{equation}
which describes the connectivity of the graph.
\subsection{Redundancy and pleiotropy functions}
We define the overall measure of redundancy for the whole network as
\begin{equation}
D=\frac{\displaystyle \sum_{i\in\mathcal{C}} O_{i}}{\norm{\mathcal{S}}},
\label{redundancyeqn}
\end{equation}
where $D$ is the redundancy, $O_{i}$ the out-degree or number of links out of client $i \in \mathcal{C}$, and $\mathcal{S}$ the set of servers.
Similarly, the overall measure of pleiotropy is
\begin{equation}
L=\frac{\displaystyle \sum_{i\in\mathcal{S}}I_{i}}{\norm{\mathcal{C}}},
\label{pleiotropyeqn}
\end{equation}
where $L$ is the pleiotropy, $I_{i}$ the in-degree or number of links into server $i \in \mathcal{S}$, and $\mathcal{C}$ the set of clients.
\section{RESULTS}
\subsection{Overview}
\begin{figure}
 \centering{\resizebox{9.1cm}{!}{\includegraphics{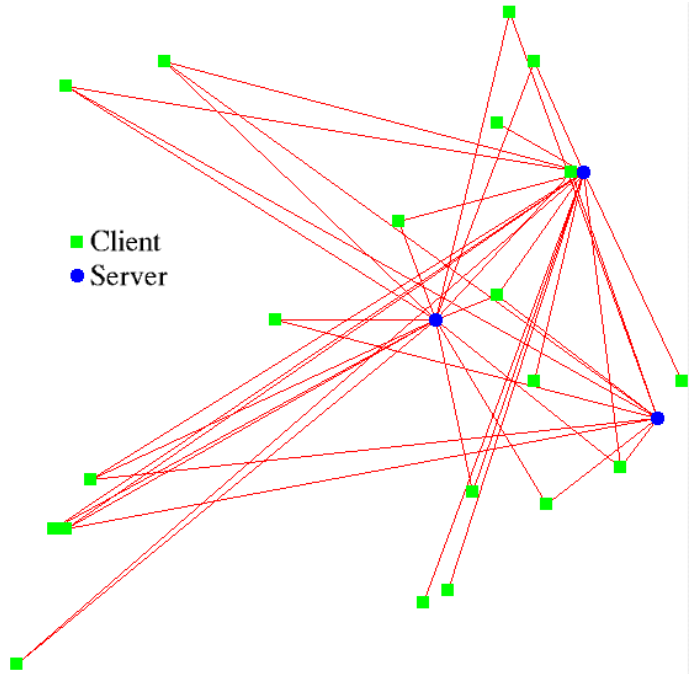}}}
  \caption{This figure shows an example of an evolved network, with clients and servers and a set of links between them. The clients and servers have been positioned at random, with a minimum spacing to avoid clutter.}
\label{network}
\end{figure}

We used both GA strategies to build a network for a number of different link failure probabilities and repair rates, and evaluated the performance of the strategies, and found the best network parameters to use. Using these network parameters, we then used the best GA strategy to find the best network possible. An example of an evolved network is shown in Figure~\ref{network}. The initial network contains no links, the first ones added with the first mutation step of the GA. 
\subsection{Varying failure probability}
\begin{figure}
\centering
  \subfigure[The fitness, pleiotropy, and redundancy measures for networks evolving with a link failure probability of 10\%.]{
  \includegraphics[width=8.1cm]{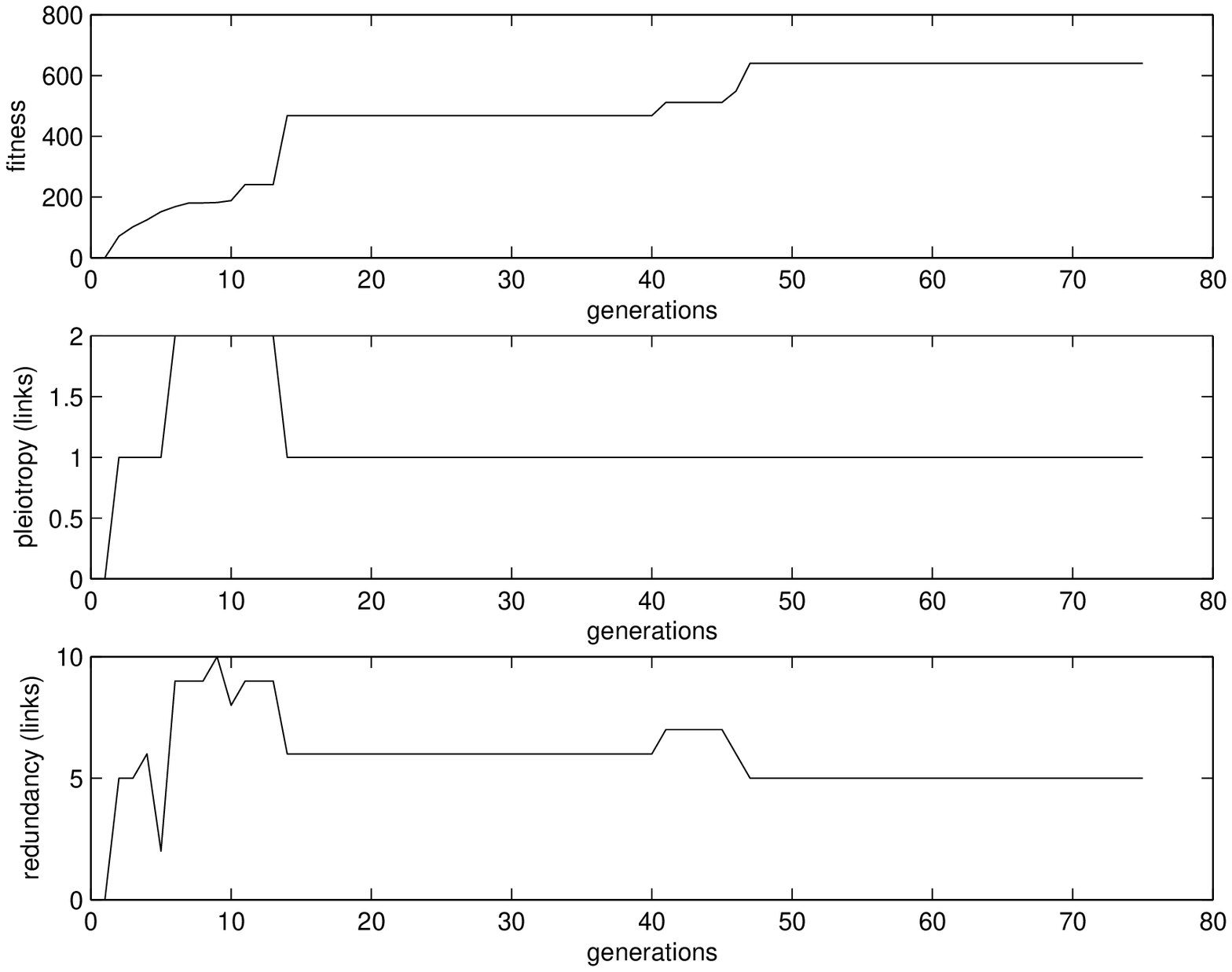}
  \label{lf10}
  }
    \subfigure[The fitness, pleiotropy, and redundancy measures for networks evolving with a link failure probability of 0.001\%.]{
  \includegraphics[width=8.1cm]{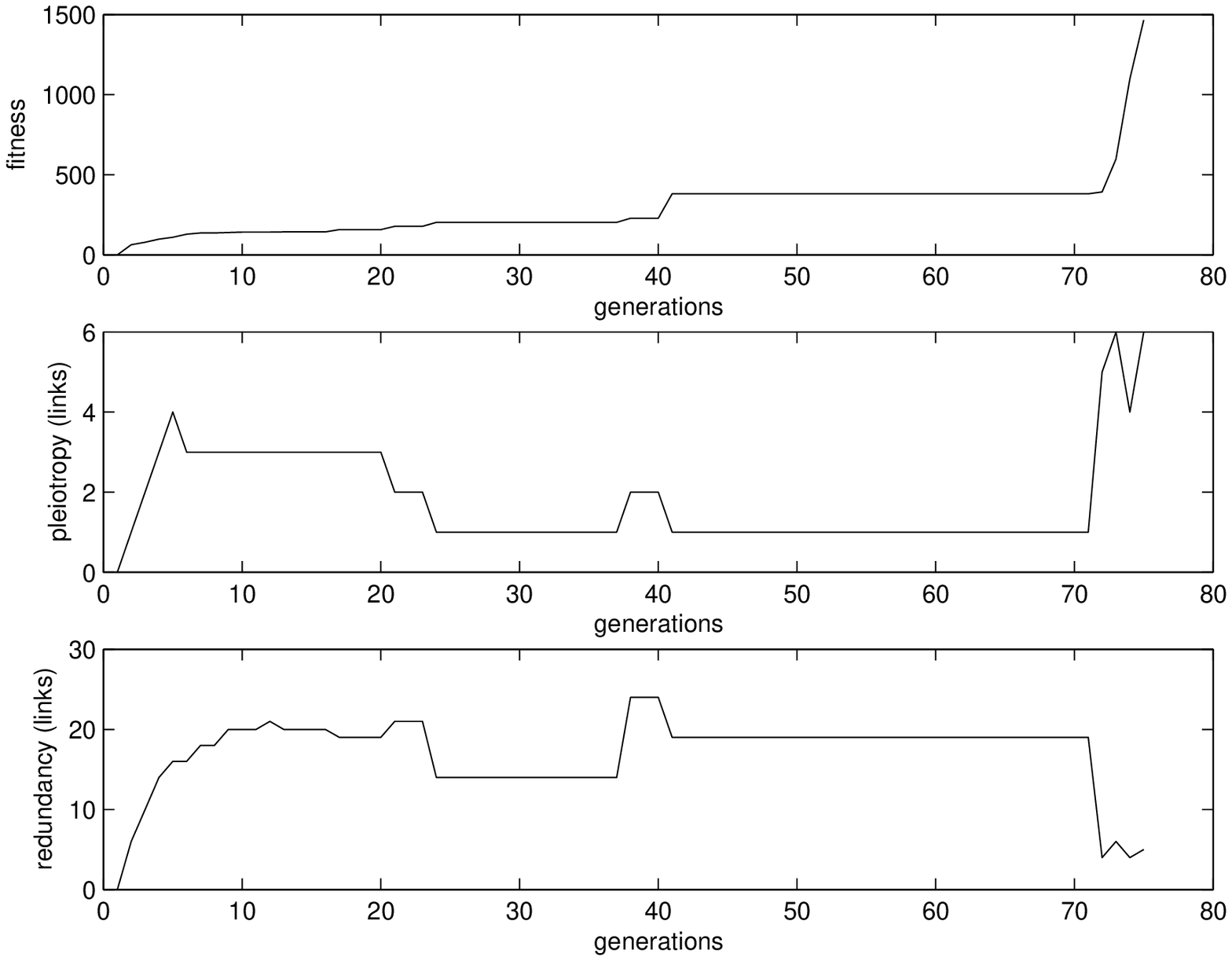}
  \label{lf0001}
  }
\caption{Here we plot the fitness, pleiotropy, and redundancy measures in of the current best network out of a set of networks that are evolving through 75 generations, where the networks are evolving to different link probabilities. As expected, a system with greater probability of link failure converges to a lower pleiotropy and higher redundancy.}
\label{figa}
\end{figure}

Here we consider the rate of convergence of the genetic algorithm, the cost, and fitness functions for varying link failure probabilities. The rate of convergence is defined as the time to three successive generations with the same maximum fitness. As a worst-case scenario, we consider a link failure probability of 10\%, and consider a range of failure probabilities as low as $0.001\%$. The results of this are shown in Table~\ref{tablea} and Figure~\ref{figa}. 
\begin{table}[htbp]
\centering
\caption{This table shows the convergence time in generations, the maximum fitness (no units), the final cost ('000s of arbitrary units), final pleiotropy (network links) and final redundancy (network links) for varying link failure probabilities. We have averaged the results over five runs of the network optimizer, considering 75 generations each run, and show the average result $\pm$ one standard deviation.}
\begin{tabular}{|c|c|c|c|c|c|}\hline
Link failure prob. & Convergence time & Max fitness & Final cost & Final pleiotropy & Final redundancy\\\hline
10\% & $14.2\pm1.6$ & $469.5\pm124.6$ & $8.9\pm2.9$ & $0.8\pm0.4$ & $9.4\pm2.6$ \\\hline
1\% & $14.0\pm3.4$ & $338.0\pm80.1$ & $10.9\pm4.1$ & $1.0\pm0.7$ & $17.2\pm2.3$\\\hline
0.1\% & $13.8\pm3.1$ & $321.7\pm159.8$ & $13.6\pm5.2$ & $1.4\pm0.9$ & $20.4\pm6.7$\\\hline
0.01\% & $12.8\pm2.1$ & $236.0\pm151.8$ & $10.0\pm6.9$ & $1.2\pm1.3$ & $18.4\pm11.7$\\\hline
0.001\% & $10.4\pm4.5$ & $528.9\pm571.9$ & $18.7\pm14.2$ & $2.8\pm3.0$ & $18.2\pm7.5$\\\hline
\end{tabular}
\label{tablea}
\end{table}
\subsection{Impact of population size on convergence to solution}
Here we consider the same parameters as for the previous subsection, but with the link failure probability set to 1\% and for varying population sizes. Again, the rate of convergence is defined as the time to three successive generations with the same maximum fitness. Not shown here is the average time to evolve through the 75 generations, which takes longer for longer populations, since more offspring have to be generated and therefore the fitness function has to be computed more often. The results of this are shown in Table~\ref{tableb}. 
\begin{table}[htbp]
\centering
\caption{This table shows the convergence time in generations, the maximum fitness (no units), the final cost (arbitrary units), final pleiotropy (network links) and final redundancy (network links) for varying population sizes. We have averaged the results over five runs of the network optimizer, considering 75 generations each run, and show the average result $\pm$ one standard deviation.}
\begin{tabular}{|c|c|c|c|c|c|}\hline
Pop.~size & Convergence time & Max fitness & Final cost & Final pleiotropy & Final redundancy\\\hline
6 & $14.0\pm1.6$ & $221.8\pm39.6$ & $15.5\pm5.5$ & $1.4\pm0.9$ & $20.8\pm5.1$\\\hline
10 & $12.8\pm3.0$ & $208.4\pm24.5$ & $18.5\pm3.2$ & $2.0\pm0.0$ & $19.2\pm2.0$\\\hline
15 & $13.6\pm2.6$ & $327.2\pm96.7$ & $11.1\pm4.0$ & $1.2\pm0.8$ & $14.4\pm4.8$\\\hline
21 & $13.8\pm2.4$ & $423.6\pm69.4$ & $8.4\pm1.6$ & $0.6\pm0.5$ & $13.0\pm7.9$\\\hline
28 & $12.8\pm4.1$ & $558.6\pm404.8$ & $9.4\pm2.0$ & $0.8\pm0.4$ & $12.2\pm5.7$\\\hline
\end{tabular}
\label{tableb}
\end{table}
\section{DISCUSSION  \& CONCLUSIONS}
The cross-over operator allows the genetic algorithm to converge much faster to solutions than the mutation operator alone, and allows it to explore a much wider range of networks in the search for a solution. Our improved cost function allows the genetic algorithm to search the space around a given reliability much more effectively, since it no longer wants to remove links to reduce the cost to zero (giving an unrealistic infinite fitness). The pleiotropy and redundancy converge to a rate of about one to two links from each client to each server. For the population ranges and link failure probabilities we considered, the convergence time shows no significant difference, although it seems to  decrease for increasing link reliability.

More work is needed to analyze the convergence time in more detail, in particular, removing some of the initial graph randomness and link failures would eliminate most of the factors contributing to the high variance in convergence time for varying population sizes. It would also be an interesting idea to benchmark the genetic algorithm on a small network for which a human could easily determine the optimal solution. We also propose optimizing for reliability and cost separately, and combining the two populations using the crossover operator.
\acknowledgments
We gratefully acknowledge funding from The University of Adelaide.
\bibliography{phd}   %>>>> bibliography data in report.bib
\bibliographystyle{spiebib}   %>>>> makes bibtex use spiebib.bst
\end{document}